\pdfoutput=1

\documentclass[11pt]{article}
\usepackage[final]{acl}
\usepackage{times}
\usepackage{latexsym}
\usepackage[T1]{fontenc}

\usepackage[utf8]{inputenc}
\usepackage{microtype}
\usepackage{inconsolata}
\usepackage{graphicx}
\usepackage{subcaption}
\usepackage{wrapfig}

\title{Linguistically Communicating Uncertainty in Patient-Facing Risk Prediction Models}

\author{Adarsa Sivaprasad \\
University of Aberdeen \\
  \texttt{a.sivaprasad.22@abdn.ac.uk} \\\And
  Ehud Reiter \\
  University of Aberdeen \\
  \texttt{e.reiter@abdn.ac.uk} \\}

\begin{document}
\maketitle
\begin{abstract}
This paper addresses the unique challenges associated with uncertainty quantification in AI models when applied to patient-facing contexts within healthcare. Unlike traditional eXplainable Artificial Intelligence (XAI) methods tailored for model developers or domain experts, additional considerations of communicating in natural language, its presentation and evaluating understandability are necessary. We identify the challenges in communication model performance, confidence, reasoning and unknown knowns using natural language in the context of risk prediction. We propose a design aimed at addressing these challenges, focusing on the specific application of in-vitro fertilisation outcome prediction.
\end{abstract}

\section{Medical Risk Communication}
Medicine has found important applications of Artificial Intelligence (AI) from its early years. In recent years, however, AI tools have become increasingly end-user-patient-facing. This poses the research question of faithful risk communication associated with AI predictions and in turn building trust in these applications.

Trust in human-AI interactions, as extensively studied in psychology and cognitive science, can be attributed to the congruence of user mental models and experience interacting with AI systems \cite{MILLER20191}. In healthcare applications involving diagnostics, regulations necessitate aligning AI models with medical professionals' knowledge and domain expertise. Current healthcare research such as project DATA2PERSON \footnote{\url{https://data2person.uvt.nl/}} explore tools for personalising decision support while still keeping doctors in the loop \cite{hommes-etal-2019-personalized}. Explainable AI and interpretable models prove valuable in this context. Another category of application is healthcare models \emph{intended for expectation management}. These are models developed on a large population study intended to understand causality or identify risk possibilities. Tools such as QRisk\footnote{\url{https://qrisk.org/index.php}} \cite{Hippisley-Coxj2099} from the National Health Service (NHS) for predicting the probability of coronary heart disease is an example which is available for public use. Despite not being medical devices, patient trust in such systems can be very important in them seeking medical attention or planning better health outcomes. This paper identifies challenges in interpretability for patients, focusing on aspects of model uncertainty.
\begin{figure}
    \centering
    \includegraphics[width=\linewidth]{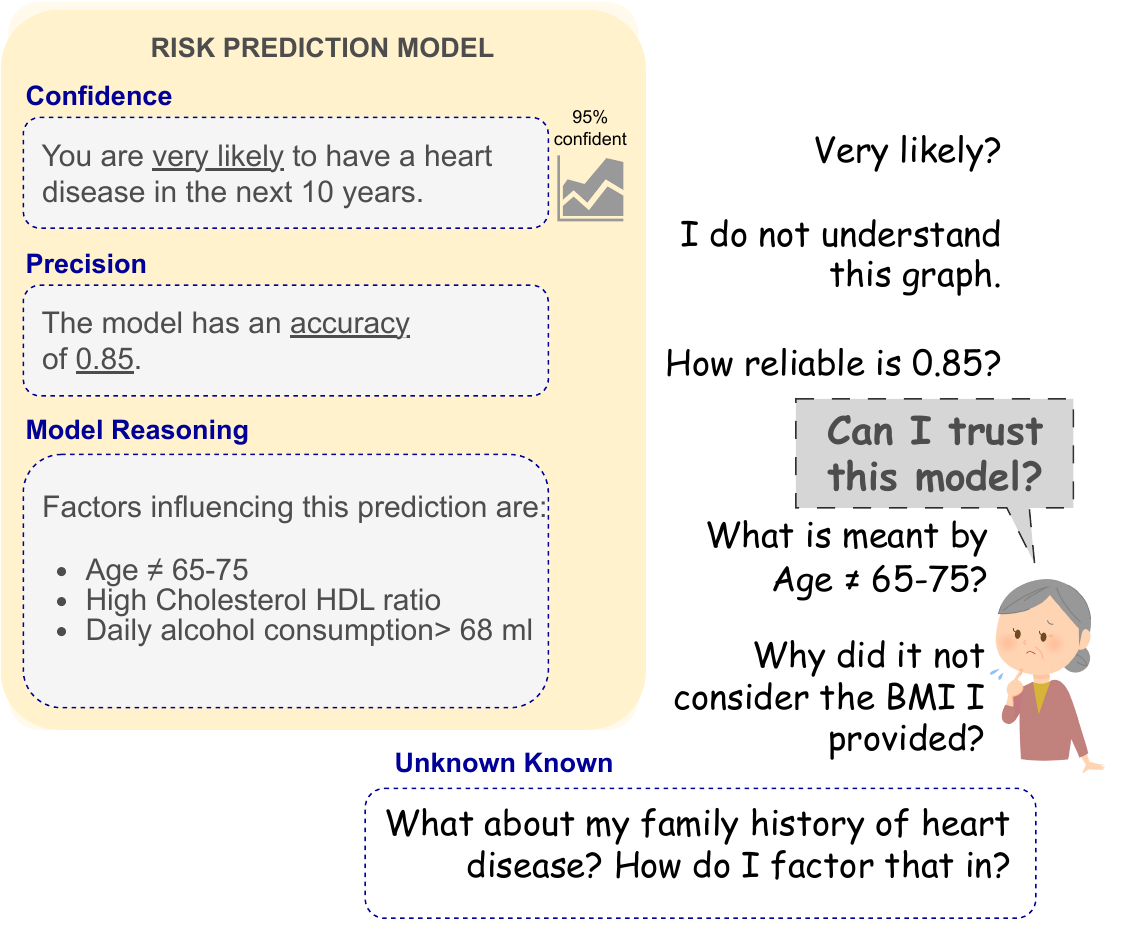}
    \caption{A patients perspective of risk communication and model explanation for a CHD prediction model.}
    \label{fig:uncertainity}
\end{figure}

Communicating uncertainty to healthcare professionals can leverage their exposure or training in understanding scientific communication, their ability to interpret probabilities, graphs and the context of their domain expertise. Public risk communication cannot make these assumptions \cite{berry2004ebook}. Population studies show variability across multiple demographic features such as differences in risk prediction comprehension based on age \cite{doi:10.1177/1071181312561002} or dependence of graph understanding on educational background \cite{doi:10.1177/0272989X11424926}. Further, there can be a complete lack, partial or fully misaligned mental models based on exposure to domain knowledge. Figure \ref{fig:uncertainity} shows different questions that a patient may seek answers to when interacting with a risk prediction model. As noted by H{\"u}llermeier et al. \cite{Hllermeier2019AleatoricAE}, in applications such as healthcare, the typical classification of uncertainty based on source is not sufficient. Further, as noted by \cite{10.1145/3461702.3462571} uncertainty can manifest as unfairness and an interdisciplinary approach that draws from literature in machine learning, visualization/HCI, design, decision-making, and fairness is required to address this.

We intend to establish the necessity for research in uncertainty communication by considering end-user needs and limitations. We illustrate this with a straightforward example of a heart disease prediction model. Our focus is to improve the user interface of the model for understandability and faithful expectation management. In section \ref{sec:IVF}, we go on to extrapolate this to a specific application of in vitro fertilisation (IVF) treatment, proposing a study design for future research.

\section{Types of Uncertainty and Current Explanation Methods}
\begin{figure}[hbt!]
    \centering
    \includegraphics[width=0.95\linewidth]{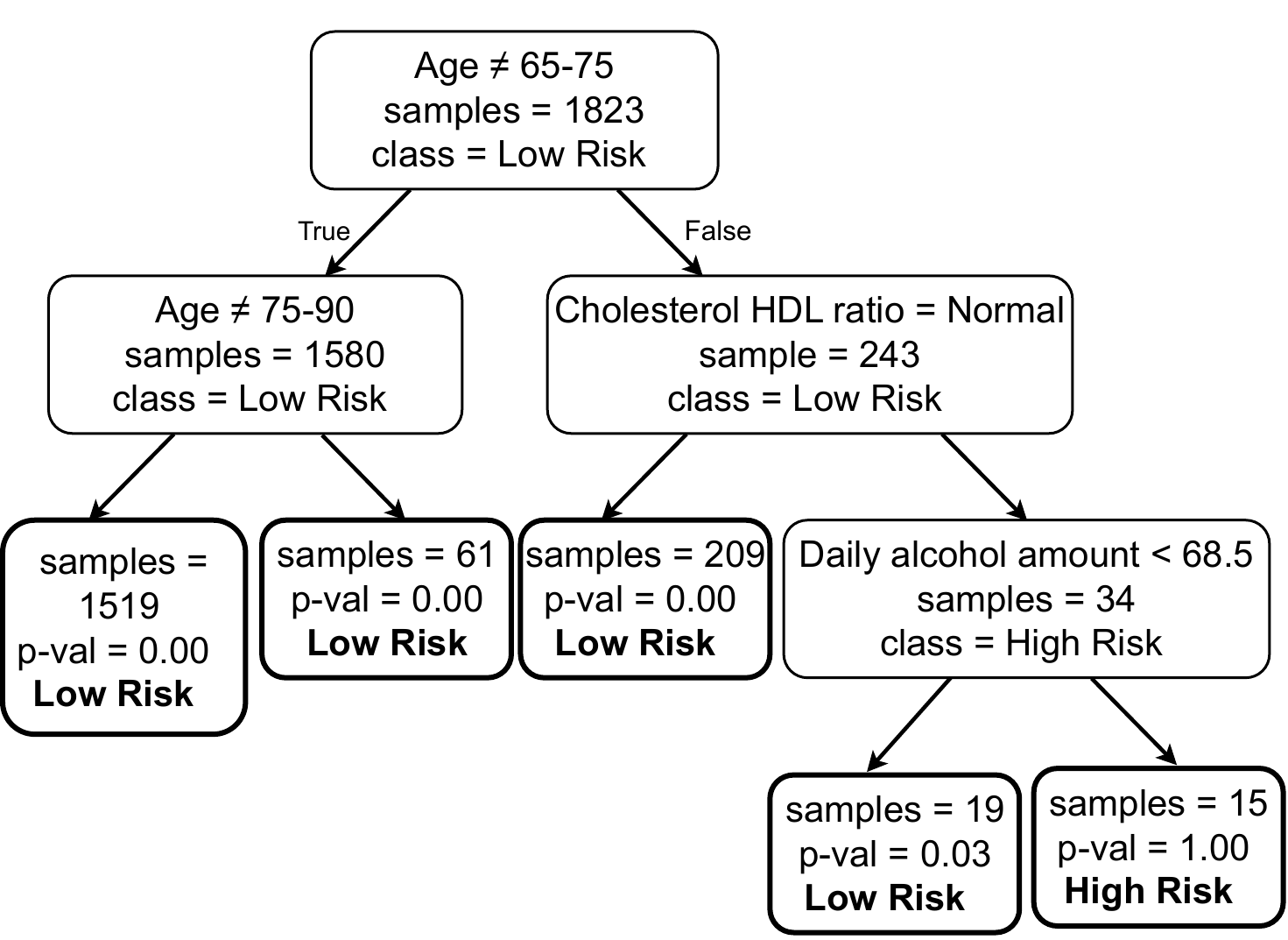}
    \caption{A CART model for predicting the risk of CHD. (Cholesterol HDL ratio -  Ratio of total cholesterol to HDL cholesterol). The number of data points corresponding to each node is denoted as \emph{samples}. Confidence of leaf node prediction based on data distribution at the node is computed using the chi-square test, and the corresponding p-value is displayed.  }
    \label{fig:DT}
\end{figure}

Before the coronavirus pandemic, coronary heart disease had been the leading cause of death worldwide for at least 30 years. Hence multiple national health organisations are invested in population studies and effectively predicting coronary heart disease (CHD). We look at the Busselton health study data subset \cite{knuiman1998multivariate} with 2279 records and 13 patient features. Whether the patient developed CHD in the next 10 years can be modelled as a binary classification problem. A simple decision tree (DT) based on the Classification and regression tree (CART) algorithm is shown in Figure \ref{fig:DT}. We use DT since they are inherently interpretable.

Consider a patient wanting to understand this classifier model. They may be seeking answers to varied questions as shown in Figure \ref{fig:uncertainity}. We look into different sources of uncertainty and measures to quantify and explain them.
\subsection{Performance Metrics}
Performance metrics serve as common indicators of model generalization, the exact metric varying based on the application or the relevant error type. In the case of CHD, based on the 456 records in test data, the model has an accuracy of 0.92 which may be considered satisfactory. Accuracy or precision is a commonly used notion in public discourse. However, here, the implication of a false negative - classifying a person with a risk of CHD as \emph{low risk}, is higher than the risk of a false positive -  classifying a person wrongly as \emph{high risk}. Since accuracy does not make this distinction, a false negative rate, false omission rate or recall is a more important heuristic even though it may be harder to interpret.

While studies have looked at the perception of expressing probability as frequency, percentages, or risk difference, Zipkin et al. \cite{zipkin2014evidence} concluded that it is neither well-understood nor popular with patients. The individual difference in perceiving probability, frequency, percentage etc \cite{Peters2008NumeracyAT}  and the impact a positive or negative framing of numbers can have on an individual's feelings \cite{Tversky1975}  have been shown in user studies. In addition to numeric representation, the use of graphical, textual and tabular risk communication methods are well explored in literature\cite{doi:10.1146/annurev-statistics-010814-020148}. There is no conclusive evidence of an appropriate method. However, as noted by Reiter et al. \cite{reiter-2019-natural} good explainable AI methods targeted towards readers should have a narrative structure.

\subsection{Confidence of a Prediction}
The CART algorithm picks a decision node based on the Gini index - a measures of how often a randomly chosen element of a set would be incorrectly labelled.  A measure of confidence at a node (confidence of a prediction) should also consider the number of training data points that follow the particular rule. A rule followed by a larger number of training data is more certain than a rule/node with fewer data points. We compute confidence as a p-value based on the chi-square test, and the value at each leaf node is shown in Figure \ref{fig:DT}. A p-value of 0 denotes the high statistical significance of predicted labels matching the distribution of observed labels at the node, or a 100\% confidence in the prediction and a value of 1 denotes no confidence. However, conveying this numerically to a patient is difficult. Mapping confidence intervals to verbal terms is the commonly employed approach in risk communication \cite{doi:10.1146/annurev-statistics-010814-020148}.

\subsection{Precision and Confidence}
While the terms precision and confidence are used interchangeably many times, as noted here, they are two different parameters. While the performance metric can be applied to the entire model, confidence concerns an individual prediction or a particular patient. For the CHD prediction model, the confidence values of nodes lie in the range of 0 to 1. For a patient with \(Age \neq 65-75\) \(and\) \(Age \neq 75-90\) prediction of low risk based on 1519 samples with a confidence of 0.0 is more certain than a low risk predicted for a patient with \(Age = 65-75\) \(and\) \emph{Cholesterol HDL ratio} \(\neq\) \emph{Normal} \(and\) \emph{Daily alcohol consumption} \(<\) \emph{68.5 ml} based on 19 samples (confidence of 0.03) even though the model accuracy is 0.92 in both cases. While multiple studies have looked at the verbal mapping of a single probability measure conveying certainty with both precision and confidence remains unexplored. Appendix \ref{appendix:1} presents a possible approach to combining the two measures. We note that a visual representation of model calibration as reliability diagrams -  a plot of expected sample accuracy as a function of confidence combines the two probabilities is another possible approach to be further explored.

\subsection{Model Reasoning}
\begin{figure}[hbt!]
    \centering
    \includegraphics[width=0.95\linewidth]{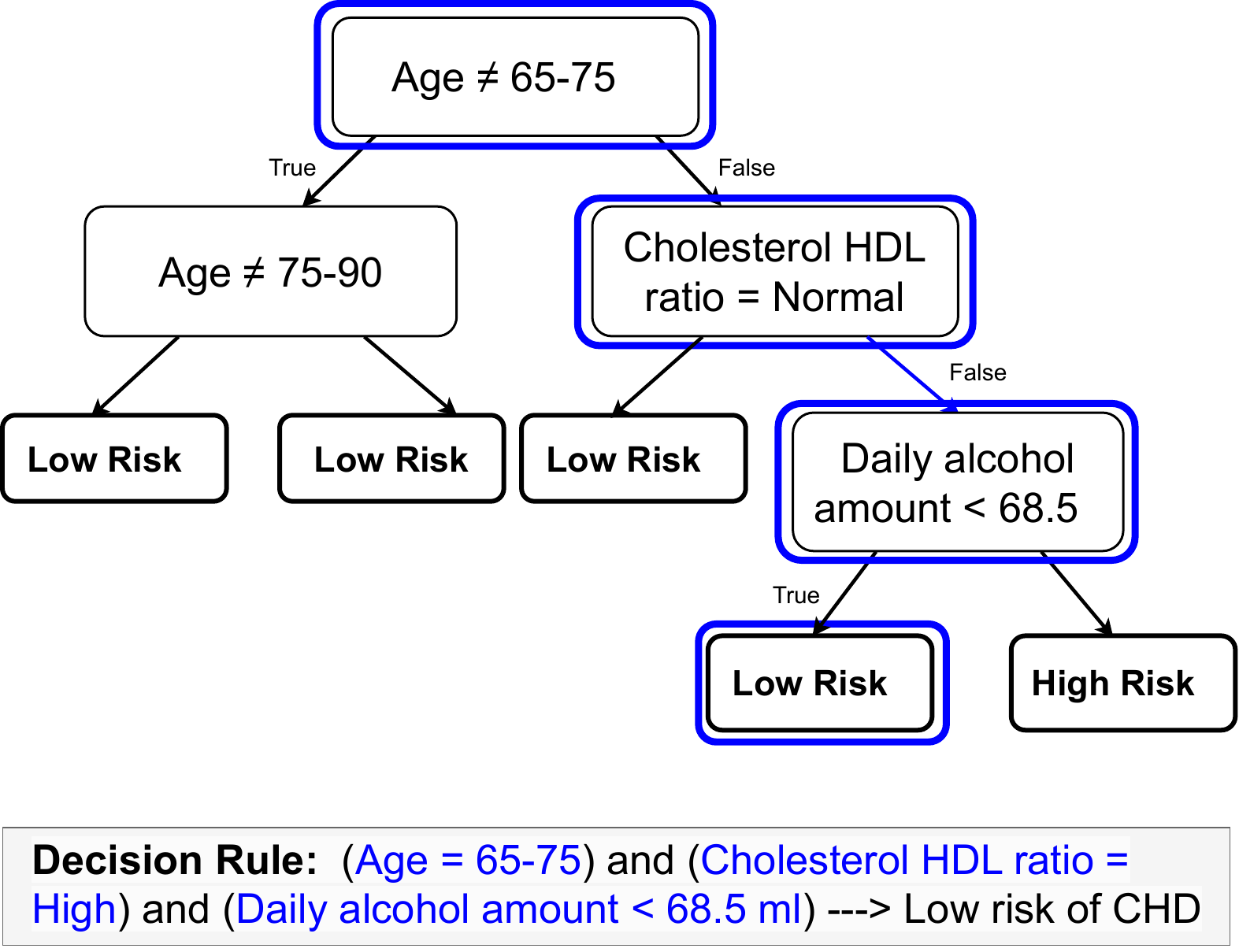}
    \caption{The decision path followed along a given DT for a particular patient input.  The model predicts \emph{low risk} following the decision path highlighted in Blue.}
    \label{fig:DTexplnation}
\end{figure}

While not usually associated with uncertainty calibration, the difference in model reasoning and mental model of the end-user patient in this case is crucial in expectation management. Model interpretation methods attempt to do this. Figure \ref{fig:DTexplnation} shows the decision path along the tree for a particular prediction and textual representation of the same. This can explain to the patient that \emph{\(Age\) not between 65-75}, \emph{Cholesterol HDL ratio being Normal} and \emph{Drinking amount <68.5 ml/day} have led to the decision. A local explanation of this manner does not provide any causal or counterfactual reasoning. It does not help in answering the question of how to alter this prediction. More importantly, it does not tell the patient, under what condition this model does not hold. As observed in \cite{sivaprasad2023evaluation}, this can be addressed using global model explanations. However global explanations also increase the complexity and cognitive load in understanding \cite{DBLP:journals/corr/abs-1902-00006}. A way around this would be to evaluate systems on the change in the mental model of the patient post-exposure to explanation. Multiple methods of eliciting user's mental model has been explored, which are summarised in \cite{10.3389/fcomp.2023.1114806} and we propose to build it into the explanation interface.

\subsection{Unknown Knowns}
Žižek et al. \cite{vzivzek2006philosophy} identify that the "disavowed beliefs, suppositions, and obscene practices we pretend not to know about, form the background of our public values". This transpires to our perception of risk too. When modelling CHD there are factors not considered in the model (\emph{BMI} is available in the dataset but not a decision node). It is also possible that a factor that is known (to the patient) as contributing to CHD - a preexisting mental model of CHD risk, is unavailable in the dataset. The Busselton dataset does not contain genetic information, but a patient who has a history of CHD in the family would be interested in knowing the effect of genetic factors on risk prediction. While explanations aim to bridge this gap \cite{MILLER20191} it need not be always possible within the scope of an available dataset and model space.  

\section{Expectation Management in Infertility Treatment }
\label{sec:IVF}
According to a World Health Organization (WHO) report in 2021, one in six individuals worldwide experiences infertility at some point in their lives. Infertility treatments not only have health implications but also exert social, psychological, and economic impacts on individuals. Therefore, an outcome prediction tool for patients at different stages of infertility treatment can be a very beneficial tool in risk assessment and expectation management for those involved. We look at publicly available Outcome Prediction in Subfertility Tool (OPIS) \footnote{\url{https://w3.abdn.ac.uk/clsm/opis/}}. This tool is built on the McLernon model \cite{mclernon2016predicting}.  The model is trained on data from 113873 women (cross-validated with a reported C index of 0.72) and validated on external data in a different geographical context and a more recent time carried out by \cite{Leijdekkers2018-sk}. OPIS contains two tools and we use the post In-vitro fertilisation(IVF) tool for the study.

\begin{figure}[h]
    \centering
    \includegraphics[width=\linewidth]{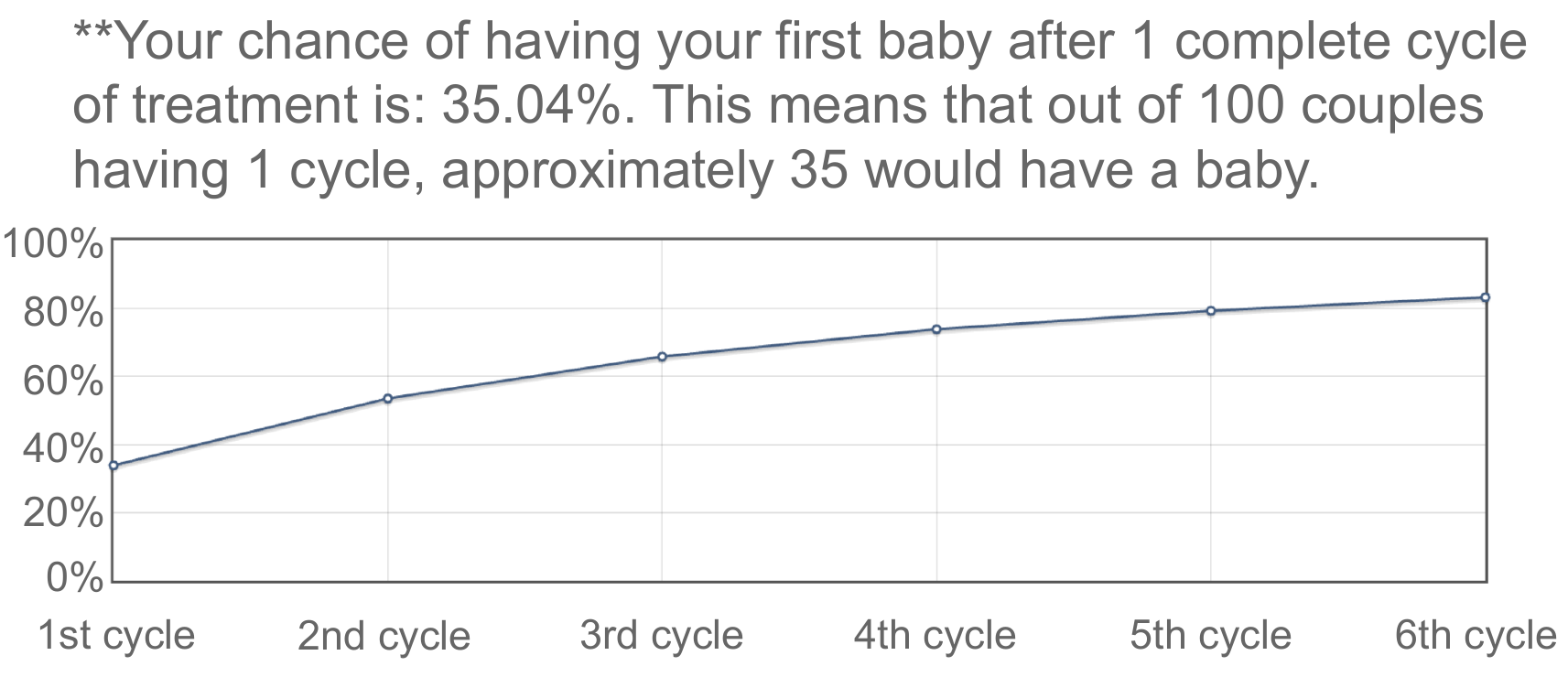}
    \caption{The cumulative probability over 6 IVF cycle  displayed as graph in the OPIS tool. Corresponding patient input - \emph{Age} = 34; \emph{Years of infertility} = 0; \emph{Number of eggs collected in first IVF cycle} = 1; \emph{Type of embryo transfer} = Stage 2 embryos transferred on day 2 or 3 ; \emph{Previous pregnancy} =  No; \emph{Tubal infertility} = No; \emph{First cycle type} = IVF; \emph{Embryos frozen in first cycle} = Yes.}
    \label{fig:opis}
\end{figure}

The user interface of the tool is shown in Figure \ref{fig:opis}. Based on the patient features input, the cumulative probability of live birth in different IVF cycles is displayed as a graph. The probabilities are computed based on a discrete-time logistic regression model.
We looked into the feedback of 44 users who used the tool between 2021-2023 \footnote{Personal communication from Dr. David McLernon, 2023}. Of this, 37 were patients and the rest identified as healthcare professionals. While all the healthcare professionals felt that the tool was user-friendly with regard to output presentation, 13\% of the patients did not think so. Further, 24\% of patients said they did not understand what their results meant. This perceived lack of understandability may cause confusion, and lack of trust and hence calls for research into the way results are communicated to the patient.

The most recurring pattern of feedback from the patients involved the question: \emph{In addition to the features input in the interface, I additionally have specific attributes that may be important. Does the model prediction still apply to me?} Putting this in the context of uncertainties and explanation methods discussed in the previous section, there is a clear need for better explanation and presentation of results. Our future work aims to address this. 

\textbf{Understanding patient expectations}:  We have established the need for understanding user expectations and designing explanations tailored to improve understandability. The currently available feedback is anonymous. It does not consider the user's ease of understanding numeric values, ability to comprehend graphs, and their existing domain knowledge. We plan to find this out through a qualitative study. Users of the tool will be asked to provide demographic information, personal characteristics \cite{EasyToPlease}, their feedback and expectations from the tool.

\textbf{Communicating uncertainty} : The current user interface presents probability as a graph and calculates the chance of a live birth in a specific cycle cumulatively from previous cycles. Building on the earlier discussion about communicating probabilities, we will also factor the confidence of predictions, and assess the effectiveness of providing a textual explanation for the chance of a live birth. This assessment will be conducted using the current user interface as the baseline.

\textbf{Model reasoning}: Currently, the tool does not provide any information about the model. Typically, explaining a regression model involves highlighting feature importance. The McLernon model considers over 20 patient features along with the time interval between cycles. We will assess the understandability, change in mental model through comprehension tasks, instead of using ratings for feedback. Additionally, we will explore alternative interpretable models like decision trees for comparison. To achieve a balance between complexity and the need for a global model explanation, we plan to use an explorative user interface. A dialogue-based system that can be navigated with a fixed set of instructions as proposed in \cite{Slack2023} would be a promising approach.

\textbf{Unknown knowns}: The OPIS tool is built on data gathered in the United Kingdom between 1999 and 2008. Since then, there have been advancements in treatment methods and understanding of patient information. The current model lacks information on factors such as patient BMI and smoking status, which are recognized as influencers of infertility. The prominent pattern in patient feedback as pointed out earlier \emph{"... I additionally have specific attributes that may be important."} directly points to the need for addressing this aspect in the model explanation. We suggest incorporating information from more recent sources. We plan to enhance explanations by integrating data from other countries and leveraging expert knowledge in the field. 

\section{Limitations}
The Busselton dataset used for coronary heart disease (CHD) prediction in this study, is relatively small, rendering the findings not fully representative of the broader CHD prediction landscape. Also, a larger dataset would result in more complex models hence posing further aspects to be considered in explanation generation. Nonetheless, the identified problem areas remain pertinent. 
An assessment of the highlighted issues within the context of CHD has not been executed with actual patients or medical professionals, potentially influencing the usefulness of model explanation examples used. This will be validated in the context of IVF prediction study, where actual patients will be recruited to provide feedback on the explanations. 
Completely addressing the challenge of \emph{unknown knowns} falls beyond the scope of this research. We acknowledge its presence and seek to draw attention to it within the broader research community.

\section{Ethical considerations}
AI developers have an ethical obligation to provide truthful and accurate explanations. As argued in \cite{wachter2018counterfactual}, building trust is essential to increase societal acceptance of algorithmic decision-making. We offer a means to achieve this in the context of risk prediction. The proposed study with IVF outcome prediction involves the evaluation of tool with patients. Hence ethical approvals from appropriate regulatory authorities will be obtained.

\section*{Acknowledgements}
We thank Dr. David McLernon for his continued support in designing the proposed study on OPIS and sharing the feedback from the OPIS tool. We would like to thank Prof. Ingrid Zukerman and Dr Sameen Maruf for generously sharing their expertise in utilizing the Busselton dataset and their discussions on uncertainty communication. We also thank the anonymous reviewers for their feedback which has improved this work. A. Sivaprasad is ESR in the NL4XAI project which has received funding from the European Union’s Horizon 2020 research and innovation programme under the Marie Skłodowska-Curie Grant Agreement No. 860621.

\bibliography{uncertainNLP}


\appendix
\section{Combining precision and accuracy}
\label{appendix:1}
\label{sec:appendix}

The terms used here are adapted from the verbal mapping of confidence proposed in \cite{doi:10.1146/annurev-statistics-010814-020148}. The CHD model has an accuracy of 0.92. We additionally introduce the term \emph{possibly} to account for accuracy values below 0.9. 
\begin{figure}[hbt!]
    \centering
    \includegraphics[width=0.95\linewidth]{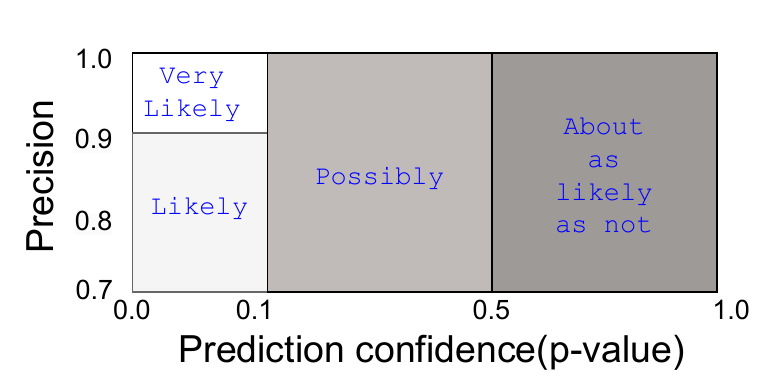}
    \caption{A verbal mapping of confidence based on precision and Gini index for CHD risk prediction. Here the range of precision and confidence scores are limited based on the model from Figure \ref{fig:DT}.} 
    \label{fig:verbalMapping}
\end{figure}
An evaluation of understandability will be explored in future studies.
\end{document}